\documentclass[9pt, conference]{IEEEtran}
\IEEEoverridecommandlockouts
\usepackage[style=numeric, sorting=none]{biblatex}
\usepackage{subcaption}
\usepackage{amsmath,amssymb,amsfonts}
\usepackage{algorithmic}
\usepackage{booktabs}
\usepackage{multirow}
\usepackage{graphicx}
\usepackage{textcomp}
\usepackage{xcolor}
\usepackage{xurl}
\def\BibTeX{{\rm B\kern-.05em{\sc i\kern-.025em b}\kern-.08em
    T\kern-.1667em\lower.7ex\hbox{E}\kern-.125emX}}
\begin{document}

\title{
SSVD: Structured SVD for Parameter-Efficient Fine-Tuning and Benchmarking under Domain Shift in ASR \\
}

\author{\IEEEauthorblockN{Pu Wang}
\IEEEauthorblockA{\textit{Department of Electrical Engineering} \\
\textit{KU Leuven}\\
Leuven, Belgium \\
pu.wang@esat.kuleuven.be}
\and
\IEEEauthorblockN{Shinji Watanabe}
\IEEEauthorblockA{\textit{Language Technologies Institute} \\
\textit{Carnegie Mellon University}\\
Pittsburgh, PA, USA \\
shinjiw@ieee.org}
\and
\IEEEauthorblockN{Hugo Van hamme}
\IEEEauthorblockA{\textit{Department of Electrical Engineering} \\
\textit{KU Leuven}\\
Leuven, Belgium \\
hugo.vanhamme@esat.kuleuven.be}
}
\maketitle

\begin{abstract}
Parameter-efficient fine-tuning (PEFT) has emerged as a scalable solution for adapting large foundation models. While low-rank adaptation (LoRA) is widely used in speech applications, its state-of-the-art variants, e.g., VeRA, DoRA, PiSSA, and SVFT, are developed mainly for language and vision tasks, with limited validation in speech. This work presents the first comprehensive integration and benchmarking of these PEFT methods within ESPnet. We further introduce structured SVD-guided (SSVD) fine-tuning, which selectively rotates input-associated right singular vectors while keeping output-associated vectors fixed to preserve semantic mappings. This design enables robust domain adaptation with minimal trainable parameters and improved efficiency. We evaluate all methods on domain-shifted speech recognition tasks, including child speech and dialectal variation, across model scales from 0.1B to 2B. All implementations are released in ESPnet to support reproducibility and future work.

\end{abstract}


\section{Introduction}
\label{sec:intro}

Large-scale foundation models pretrained on multilingual speech corpora, such as OpenAI's Whisper~\cite{radford2023robust}, the open-source Whisper-style OWSM~\cite{peng2023reproducing, peng2024owsm}, NVIDIA’s Canary~\cite{puvvada2024less}, etc., have demonstrated strong generalization and adaptability across diverse speech recognition tasks. Fine-tuning these models for downstream or personalized applications has therefore become the dominant paradigm. However, these models are typically trained on broad yet generic datasets dominated by standard accents and adult speech, limiting their effectiveness under domain shifts involving regional accents, dialectal variation, child speech, and other low-resource speech domains. In such scenarios, mismatches in acoustic features, linguistic structures, and articulatory behaviors can significantly degrade performance.

Recent studies on scaling laws of large-scale multilingual automatic speech recognition (ASR) models have shown that increasing model size substantially improves recognition accuracy, particularly for low-resource languages~\cite{chen2025owls, fan2024benchmarking, wang2023benefits, getman2024exploring}. Nonetheless, rapid scaling introduces practical challenges, as adapting large models to specific downstream or out-of-domain scenarios can become prohibitively expensive in terms of computational resources and storage requirements.

To address the limitations of full-model fine-tuning, parameter-efficient fine-tuning (PEFT) methods have been actively explored, particularly with large language models (LLMs)~\cite{houlsby2019parameter, lester2021power, zaken2022bitfit, hu2022lora}. Among them, low-rank adaptation (LoRA) has received considerable attention for significantly reducing the number of trainable parameters without modifying the original model architecture or introducing additional decoding latency~\cite{hu2022lora, gu2023qlora, zhang2023adaptive, dighe2024leveraging}. 
Specifically, LoRA freezes the pre-trained weight matrices and injects two trainable low-rank matrices to approximate weight updates. While originally proposed for LLMs, LoRA has also attracted increasing interest in speech tasks and has been successfully applied in various speech applications~\cite{dighe2024leveraging, baby2024robust, wang2022bottleneck, liu2024sparsely, xu2024towards, song2024lora}. Wang et al.~\cite{wang2022bottleneck} explore low-rankness in speech models and adapt it to spoken language understanding tasks. Liu et al.~\cite{liu2024sparsely} adapt Whisper with LoRA to child speech ASR. Xu et al.~\cite{xu2024towards} further investigate using LoRA to mitigate forgetting when fine-tuning Whisper to multiple languages. Song et al.~\cite{song2024lora} propose LoRA with a mixture of experts (MoE) to build an extensible multilingual LoRA-Whisper ASR model.

Although LoRA effectively reduces parameter count, it still incurs noticeable storage overhead as foundation models scale and suffers from convergence inefficiencies~\cite{kopiczko2024vera, meng2024pissa}. To mitigate these issues, vector-based random matrix adaptation (VeRA) has been recently proposed as a more lightweight alternative~\cite{kopiczko2024vera}. VeRA randomly initializes and freezes the low-rank matrices of LoRA, sharing these matrices across layers while updating only two scaling vectors. Thus, further reducing the memory footprint with acceptable performance drops on natural language processing (NLP) and vision tasks. Additionally, recent studies in NLP have shown a persistent performance gap between LoRA and full fine-tuning. Liu et al.~\cite{liu2024dora} analyzed the dynamic differences in weights updated between LoRA-tuned and fully fine-tuned models and proposed weight-decomposed LoRA (DoRA). DoRA reparameterizes LoRA into separate directional and magnitude components, demonstrating that tuning primarily one component is sufficient to match full fine-tuning performance in downstream NLP and vision tasks.

To further improve the training efficiency of LoRA variants, Meng et al.~\cite{meng2024pissa} and Lingam et al.~\cite{lingam2024svft} introduce singular value decomposition (SVD) to guide parameter tuning along more meaningful directions. Instead of relying on random initialization within LoRA, Meng et al.~\cite{meng2024pissa} proposed PiSSA, which initializes the low-rank matrices of LoRA using the principal singular vectors and values derived from the pre-trained weight matrix. It accelerates convergence and reduces the risk of becoming trapped in suboptimal local minima compared to standard LoRA in NLP tasks. Lingam et al.~\cite{lingam2024svft} extended this idea by proposing singular vector guided fine-tuning (SVFT), which further minimizes the number of trainable parameters by freezing the singular vectors entirely and updating only the singular values, significantly reducing parameter counts compared to PiSSA in NLP and vision domains.
 
Despite standard LoRA showing promising results in ASR, its advanced variants, as mentioned earlier, are primarily developed and studied in natural language and vision tasks, with their application to speech recognition remaining relatively underexplored.

Speech fundamentally differs from text in its dynamic, and highly variable nature, with significant acoustic diversity, pronunciation variability, dialectal shifts, etc. For instance,  Flemish (Belgian Dutch) differs from standard Dutch, with a prominent distinction being the use of the \textit{soft $/g/$}, compared to the \textit{hard $/g/$} commonly spoken in the Netherlands. Even identical lexical content ``Goed geregend, zeg!'', such phonetic differences introduce substantial acoustic mismatch. Similarly, child speech presents unique acoustic characteristics, such as phonetic variations arising from developmental articulation patterns (e.g., pronouncing ``wabbit'' instead of ``rabbit''). PEFT methods are thus expected to efficiently adapt to domain shifts in acoustic variation while maintaining consistent semantic recognition.

To address this gap and facilitate deeper understanding and advancement in PEFT specifically for speech, we provide a thorough investigation of LoRA and its leading variants. We implement and comprehensively evaluate these methods on challenging speech recognition scenarios, including low-resource child speech, as well as dialectal variations between Dutch and Flemish, by fine-tuning on OWSM models at multiple scales (from 0.1B to 2B) using the widely adopted open-source ESPnet toolkit. All implemented methods are publicly released and fully integrated into ESPnet toolkit, facilitating reproducibility and enabling further research in speech-specific PEFT.

Meanwhile, this paper introduces a novel structured SVD-guided (SSVD) PEFT approach specifically designed to handle domain shifts in speech recognition tasks. The proposed SSVD method leverages SVD to first decompose pre-trained model weights into right singular vectors, representing the input (acoustic) feature space, and left singular vectors, capturing the output (semantic) feature space. During fine-tuning, our method selectively applies structured rotations to the input-associated right singular vectors, effectively adapting acoustic feature representations to align better with domain-shifted inputs. Meanwhile, the left singular vectors remain fixed, preserving stable semantic mappings. This innovative approach ensures efficient adaptation to acoustic domain shifts while maintaining robust linguistic outputs.

The main contributions of this paper are:

\begin{itemize}
\item Providing the first comprehensive benchmarking of SoTA PEFT methods, including LoRA, VeRA, DoRA, PiSSA, and SVFT, within the widely adopted ESPnet framework~\cite{watanabe2018espnet}. These methods are comprehensively evaluated on challenging speech recognition tasks involving low-resource child speech and dialectal variations (Dutch vs. Flemish) at multiple model scales, making all implementations publicly available and fully integrated into ESPnet for reproducibility and future research.
\item Introducing a novel structured SVD-guided (SSVD) PEFT method that explicitly designed to efficiently adapt large-scale speech foundation models to challenging domain shifts by selectively adapting input-related singular vectors, while preserving semantic mappings through fixed output-related singular vector components.
\item Empirically demonstrating that SSVD achieves comparable performance with significantly fewer trainable parameters and higher efficiency than LoRA and SoTA LoRA variants, approaching fully fine-tuned model performance.
\end{itemize}

\section{Related Work for Benchmarking}
\label{sec:relate_work}
\begin{table*}[bhtp]
    \centering
    \caption{Summary of PEFT methods formulations and \underline{trainable parameter} counts (\underline{underlined variables}).}
    \label{tab:comparison_methods}
    \renewcommand{\arraystretch}{1.8}
    \resizebox{0.9\textwidth}{!}{
    \begin{tabular}{cccccc}
        \hline\hline
        \textbf{LoRA}~\cite{hu2022lora} & \textbf{VeRA}~\cite{kopiczko2024vera} & \textbf{DoRA}~\cite{liu2024dora} & \textbf{PiSSA}~\cite{meng2024pissa} & \textbf{SVFT}~\cite{lingam2024svft} & \textbf{SSVD} (Ours) \\
        \hline
        $\mathbf{W}_0 + \underline{\mathbf{A B}}^\top$~(\ref{eq:lora})&
        $\mathbf{W}_0 + ({\underline{\boldsymbol\Lambda}_{d}}\mathbf{A})({\underline{\boldsymbol\Lambda}_{b}}\mathbf{B})^\top$~(\ref{eq:vera})&
        $\displaystyle \underline{\mathbf{m}} \frac{\mathbf{W}_0 + \underline{\mathbf{A B}}^\top}{\|\mathbf{W}_0 + \underline{\mathbf{A B}}^\top\|_c}$~(\ref{eq:dora})&
        $\underline{\mathbf{A}\mathbf{B}}^\top + \sum_{i=k+1}^{n-k} \sigma_i \mathbf{u}_i \mathbf{v}_i^\top$~(\ref{eq:pissa}) &
        $\mathbf{U}(\mathbf{\Sigma}+\underline{\mathbf{M}})\mathbf{V}^{\top}$~(\ref{eq:svft}) &
        $\mathbf{U} \boldsymbol({\Sigma + \underline{\mathbf{\Delta \Sigma}}})\underline{\mathbf{G}}\mathbf{V}^{\top}$~(\ref{eq:ssvd})\\
        $r \times (m+n)$ & $r+m+1$ & $r \times (m+n)+m$ & $r \times (m+n)$ & $n \times q +(n-q)(q+1)$ & $\displaystyle \frac{k(k+1)}{2}$\\ 
        \hline\hline
    \end{tabular}
    }
\end{table*}
In this section, we introduce the state-of-the-art (SoTA) LoRA variants evaluated in this work. Table~\ref{tab:comparison_methods} summarizes their corresponding formulations and trainable parameter counts. \underline{Trainable variables} in this study are indicated by \underline{underlining}.

\textbf{LoRA}. LoRA freezes the pre-trained weights of models and injects trainable low-rank matrices into each linear layer. For a pre-trained weight $\mathbf W_0\in\mathbb{R}^{m\times n}$, LoRA parameterizes its fine-tuning update as: 
\begin{equation}
\label{eq:lora}
\mathbf{W}' = \mathbf{W}_0 + \underline{\mathbf{A}\mathbf{B}}^\top,
\end{equation}
where $\mathbf A\in\mathbb{R}^{m\times r}$, $\mathbf B\in\mathbb{R}^{n\times r}$ are trainable low-rank matrices.
LoRA significantly reduces the number of trainable parameters to $r \times (m+n)$ by choosing a rank $r\ll \min(m,n)$, but as model size scales, it can still accumulate a non-trivial number of learnable parameters.

\textbf{VERA}. Unlike LoRA, which introduces trainable low-rank matrices for each adapted layer, VeRA employs a single pair of frozen, randomly initialized low-rank matrices shared across all layers. Adaptation is achieved through the learning of small, trainable scaling vectors specific to each layer. For a pre-trained weight $\mathbf W_0$, VERA adapt it as:
\begin{equation}
\label{eq:vera}
\mathbf{W}' = \mathbf{W}_0 + ({\underline{\boldsymbol\Lambda}_{d}}\mathbf{A})({\underline{\boldsymbol\Lambda}_{b}}\mathbf{B})^\top
\end{equation}
where $\mathbf A$ and $\mathbf B$ are low-rank random matrices, generated once and kept frozen. ${\boldsymbol\Lambda}_{d}$ and ${\boldsymbol\Lambda}_{b}$ are diagonal matrices formed from small trainable scaling vectors $b\in\mathbb{R}^{r}$ and $d\in\mathbb{R}^{m}$, respectively. Since $\mathbf{A}$ and $\mathbf{B}$ are shared and reproducible from a fixed random seed, VeRA significantly lowers the memory footprint to $r+m$.

\textbf{DoRA}. While VeRA focuses on achieving decent performance with minimal trainable parameters, DoRA aims to bridge the performance gap between LoRA and full fine-tuning.  For a pre-trained weight matrix $\mathbf W_0$, DoRA decomposes it into a magnitude vector $\mathbf{m}=\|\mathbf{W}_0\|_c$, where $\|.\|_c$ is the vector-wise (column-wise) norm, and a directional matrix $\displaystyle \frac{\mathbf{W}_0}{\|\mathbf{W}_0\|_c}$. By analyzing the change in the components during LoRA tuning versus full fine-tuning, Liu et al.~\cite{liu2024dora} observed that LoRA struggles to simultaneously learn both direction and scale, while full fine-tuning tends to adapt primarily one of the two. DoRA hereby simplifies optimization and focuses solely on directional adaptation, injecting low-rank updates only into the directional component: 
\begin{equation}
\label{eq:dora}
\mathbf{W}' = \mathbf{\underline{m}} \frac{\mathbf{W_0}+\mathbf{\underline{AB}}^\top}{\|\mathbf{W_0}+\mathbf{\underline{AB}}^\top\|_c}
\end{equation}
Compared to LoRA, DoRA introduces an additional trainable magnitude vector $\mathbf{m} \in\mathbb{R}^{m}$, increasing the trainable parameter count to $r\times(m+n)+m$. However, this increase remains negligible compared to the size of the full model.

\textbf{PiSSA}. Unlike LoRA, VeRA, and DoRA, which rely on randomly initialized low-rank matrices and suffer from inefficient early training or suboptimal local minima, PiSSA initializes the low-rank components $\mathbf{A}$ and $\mathbf{B}$ using the top-$k$ principal singular vectors and values from the SVD of the pre-trained weight matrix $\mathbf{W}_0\in\mathbb{R}^{m\times n}$. 

Given the SVD, with $m \geq n$ (e.g., a feedforward layer):
\begin{equation}
\label{eq:svd}
\mathbf{W}_0 = \mathbf{U} \boldsymbol{\Sigma} \mathbf{V}^{\top} = \sum_{i=1}^{n} \sigma_i \mathbf{u}_i \mathbf{v}_i^\top,
\end{equation}
where, $\mathbf U\in\mathbb{R}^{m\times n}$, $\mathbf V\in\mathbb{R}^{n\times n}$ contain the left and right singular vectors $\mathbf{u}_i$ and $\mathbf{v}_i$ respectively, and $\boldsymbol{\Sigma}\in\mathbb{R}^{n\times n}$ is a diagonal matrix of singular values $\sigma_i$. 
Fine-tuning is performed along the $k$ principal directions, while the residual components with the $(n-k)$ smallest singular values are frozen:
\begin{equation}
\label{eq:pissa}
\mathbf{W}' =  \underline{\mathbf{A}\mathbf{B}}^\top + \sum_{i=k+1}^{n-k} \sigma_i \mathbf{u}_i \mathbf{v}_i^\top,
\end{equation}
Here, $\mathbf{A}$ and $\mathbf{B}$ are initialized by $\sum_{i=1}^{k} \sigma_i^{\frac{1}{2}} \mathbf{u}_i$ resp. $\sum_{i=1}^{k} \sigma_i^{\frac{1}{2}} \mathbf{v}_i^{\mathrm T}$.
This initialization enables more efficient optimization by aligning parameter updates with meaningful directions in the model's parameter space, leading to faster convergence and improved performance compared to standard LoRA.

\textbf{SVFT}. SVFT extends PiSSA by further reducing the number of trainable parameters, modifying only a sparse subset of singular values and associated directions. Starting from the SVD of a pre-trained weight matrix~(\ref{eq:svd}),
SVFT updates the weight as:
\begin{equation}
\label{eq:svft}
\mathbf{W}' = \mathbf{U}(\mathbf{\Sigma}+\underline{\mathbf{M}})\mathbf{V}^{\top} 
\end{equation}
where $\mathbf{M}\in\mathbb{R}^{n\times n}$ is a small, learnable matrix. Different structures of $\mathbf{M}$ lead to different SVFT variants:
1) $SVFT^p$ uses a diagonal matrix $\mathbf{M}$, adapting only singular values (akin to reweighting frozen singular directions);
2) $SVFT_d^B$ uses a banded matrix to introduce learnable off-diagonal interactions;
3) $SVFT_d^R$ samples $\mathbf{M}$ as a fixed random matrix;
4) $SVFT_d^T$ makes the top-$k$ strong interactions between singular vector directions learnable.
These variants enable SVFT to explicitly adjust the most critical directions in the parameter space. Among them, $SVFT_d^B$ demonstrated the best performance in~\cite{lingam2024svft}, and has a trainable parameter count of $n\times q +(n-q)(q+1)$, where $q$ denotes the bandwidth.

Although these methods have been extensively studied in text and vision domains, their effectiveness in speech remains largely unexplored. Moreover, none of them is fundamentally tailored to the unique characteristics of speech signals. This limitation is particularly evident in SVD-guided methods, which explicitly decompose the input and output feature spaces—an operation linked to the acoustic and semantic properties of speech.

In the following section, we highlight this intrinsic nature of SVD-based methods and introduce structured SVD-Guided fine-tuning (SSVD), which explicitly adapts the input feature space to enable efficient and robust domain adaptation in speech.


\section{SSVD Method}
\label{sec:method}

As SVD is mathematically demonstrated in Section~\ref{sec:relate_work} - \textbf{PiSSA}~(\ref{eq:svd}), each right singular vector $\mathbf{v}_i$ is mapped to the corresponding left singular vector $\mathbf{u}_i$, scaled by the singular value $\boldsymbol{\Sigma}$. 

An input $\mathbf x\in\mathbb{R}^n$, represented in the coordinate system spanned by the right singular vector $\mathbf v_i$, is mapped to an output $\mathbf y\in\operatorname{Col}\mathbf{O}\subset\mathbb{R}^{m}$ in the coordinate system spanned by the left singular vector $\mathbf u_i$. 
\begin{equation}
\mathbf{y} = \mathbf{U} \boldsymbol{\Sigma} \mathbf{V}^{\top}\mathbf{x} 
\end{equation}

Under domain shift in the speech input space, the input $\mathbf x\in\mathbb{R}^n$ is no longer aligned with the original right singular basis $\mathbf v_i$, but instead aligned with a shifted basis $\mathbf v_i'$. This constitutes an ``inner'' transform, whereas the output semantic space, governed by $\mathbf u_i$, remains unchanged (the ``outer'' transform). The shift in coordinate basis can be modeled through a rotation and scaling of the original input space. Accordingly, the transformation becomes:
\begin{equation}
\label{eq:ssvd}
\mathbf{y} = \mathbf{U} \boldsymbol({\Sigma + \underline{\mathbf{\Delta \Sigma}}})\underline{\mathbf{G}}\mathbf{V}^{\top}\mathbf{x}\\
           = \mathbf{W}'\mathbf{x}
\end{equation}
where, the diagonal matrix $\mathbf{\Delta \Sigma}$ models axis-wise scaling (i.e., singular value shifts), and $\mathbf{G}$ is an orthogonal matrix representing a series of rotations in the right singular vector space.

Since the principal singular values and vectors for the best matrix approximation (Eckart-Young theorem)~\cite{eckart1936approximation}, we use 
\begin{equation}
\label{eq:ssvd-k}
\mathbf{\Delta \Sigma} = \begin{bmatrix} \mathbf{\Delta \Sigma}_k & 0 \\ 0 & 0\end{bmatrix}~~\text{and}~~\mathbf{G} = \begin{bmatrix} \mathbf{G}_k & 0 \\ 0 & I \end{bmatrix}
\end{equation}
to apply adaptations to the top-$k$ components. In Section~\ref{sec:results-asr}, we discuss different choices for the proportion of adapted components.


To implement rotation within the inner transformation, we explore three parameterizations of the transformation matrix $\mathbf{G}_k$, which trade off orthogonality constraints for computational efficiency.

\subsection{Strict orthogonal constraint}
\label{sec:method-strict}
We enforce strict orthogonality by defining $\mathbf{G}_k$ using the Cayley transform~\cite{trockmanorthogonalizing}:
\begin{equation}
\mathbf{G}_k = (\mathbf{I} - \mathbf{K})(\mathbf{I} + \mathbf{K})^{-1},
\end{equation}
where $\mathbf{K}\in\mathbb{R}^{k\times k}$ is a skew-symmetric matrix (i.e, $\mathbf{K}^T = -\mathbf{K}$). By learning a skew-symmetric matrix $\mathbf{K}$, the number of trainable parameters in SSVD is reduced to $(\frac{k(k-1)}{2}+k)$, which is significantly fewer than the $(k^2+k)$ parameters required to directly learn a full matrix $\mathbf{G}_k$ (`$+k$' accounts for singular values scaling update).

However, the matrix inversion in the Cayley transform incurs a computational cost of approximately $\mathcal{O}(k^3)$ cost, which becomes prohibitive for large dimensions $k$. To address this, we introduce an approximate orthogonality constraint using a first–order Cayley approximation in~\ref{sec:method-approximate}.

\subsection{Approximate orthogonal constraint}
\label{sec:method-approximate}
When \(\lVert\mathbf K\rVert\! \ll\! 1\), we can approximate the inverse in the Cayley transform via a truncated Neumann series:
\begingroup
\setlength{\abovedisplayskip}{0.5em}
\setlength{\belowdisplayskip}{0.5em}
\setlength{\abovedisplayshortskip}{0pt}
\setlength{\belowdisplayshortskip}{0pt}
\begin{equation}
(\mathbf I+\mathbf K)^{-1} = \mathbf I-\mathbf K + \mathcal O(\mathbf K^{2})
\end{equation}
\endgroup
leading to a first-order approximation:
\begingroup
\setlength{\abovedisplayskip}{0.5em}
\setlength{\belowdisplayskip}{0.5em}
\setlength{\abovedisplayshortskip}{0pt}
\setlength{\belowdisplayshortskip}{0pt}
\begin{equation}
    \mathbf{G}_k
    \;\approx\;
    (\mathbf I-\mathbf K)(\mathbf I-\mathbf K)
    = \mathbf I - 2\mathbf K + \mathcal O(\mathbf K^{2}).
\end{equation}
\endgroup

By keeping only the linear term, we obtain the simplified form:
\begingroup
\setlength{\abovedisplayskip}{0.5em}
\setlength{\belowdisplayskip}{0.5em}
\setlength{\abovedisplayshortskip}{0pt}
\setlength{\belowdisplayshortskip}{0pt}
\begin{equation}
\mathbf G_k \approx \mathbf I - 2\mathbf K\,
\end{equation}
\endgroup
which avoids matrix inversion and reduces the cost to \(\mathcal O(k^{2})\). This approximation introduces an orthogonality error of order \(\lVert\mathbf K\rVert^{2}\), which is often acceptable for small $k$, a common setting in PEFT. We use \textbf{approximate orthogonality constraint} by default in this study. In Section~\ref{sec:results-alba}, we compare the strict and approximate orthogonality settings and demonstrate that the resulting error is trivial.

\subsection{Unconstrained rotations}
\label{sec:method-none}

As a further relaxation, we drop the orthogonality constraint altogether, allowing $\mathbf{G}_k\in\mathbb{R}^{k\times k}$
to be a freely parameterized matrix. This maximally reduces computational overhead but sacrifices the beneficial geometric properties of orthogonal transformations, such as preserving norms and angles. It also increases the number of trainable parameters to $(k^2 + k)$. The performance of this unconstrained rotation is also evaluated in Section~\ref{sec:results-alba}.

\section{Experiments}
\label{sec:experiment}

In this study, we choose the open Whisper-style speech models OWSM~\cite{peng2023reproducing, peng2024owsm, chen2025owls} as our initial models, as they support a wide range of model scales, from 0.1B to 18B parameters. Moreover, the OWSM series is fully open-sourced, providing researchers with detailed information about the corpora used in pre-trainig. This transparency ensures that there is no domain overlap in our evaluation setup, allowing for a reliable study of domain-shift issues in ASR.

All PEFT methods are implemented within the ESPnet framework, we evaluate them by fine-tuning OWSM models at 0.1B, 1B, and 2B (OWLS) scales on two domain-shifted speech datasets: MyST~\cite{pradhan2024myst} (child speech) and CGN~\cite{oostdijk2000cgn} (dialectal Flemish and Dutch).

\subsection{Dataset}
\vspace{-1.2em}
\label{sec:experiment-data}
\begin{table}[bhtp]
    \centering
    \caption{Summary of corpora.}
    \label{tab:data_statistc}
    \renewcommand{\arraystretch}{1.2}
    \resizebox{0.43\textwidth}{!}{
    \begin{tabular}{ccccc}
        \hline\hline
        \textbf{Corpus} & \textbf{Duration (Hours)} & \textbf{Train Utts} & \textbf{Dev Utts} & \textbf{Test Utts} \\
        \hline
        MyST & $179$ & $55,703$ & $9,047$ & $10,328$ \\
        CGN &
        $341$ & $179,440$ & $25,766$ & $51,615$ \\
        \hline\hline
    \end{tabular}
    }
\end{table}
The MyST corpus~\cite{pradhan2024myst} consists of English dialogues between elementary school students and virtual science tutors, spanning eight topics. The transcriptions provide verbatim orthographic annotations that capture hesitations, repetitions, and disfluencies. Following the protocol of~\cite{attia2024kidwhisper}, we filter out utterances with a word error rate (WER) higher than 50\% on Whisper-large-v2 to ensure transcription quality.

The CGN (Spoken Dutch Corpus)~\cite{oostdijk2000cgn} is a manually annotated speech database containing approximately 900 hours of Dutch speech, including 270 hours of Flemish Dutch. It consists of 15 components covering various speaking styles, such as read speech, interviews, spontaneous conversations, and telephone dialogues. In this study, we use a subset containing both Flemish and Dutch dialects, excluding component $\mathtt c$ (spontaneous interview), component $\mathtt d$ (discussions), and component $\mathtt f$ (spontaneous telephone dialogues), resulting in a total of approximately 341 hours. 

The dataset statistics are summarized in Table~\ref{tab:data_statistc}.


\subsection{Implementation details}
\label{sec:experiment-implementation}
\vspace{-0.5em}
\begin{table}[bhtp]
    \centering
    \caption{Architectural configurations of models.}
    \label{tab:model_statistc}
    \renewcommand{\arraystretch}{1.2}
    \resizebox{0.46\textwidth}{!}{
    \begin{tabular}{ccccccc}
        \hline\hline
        \textbf{Model} & \textbf{$\#$ Params} & \textbf{Encoders} & \textbf{Decoders} & \textbf{$\#$ dim} & \textbf{$\#$ head}\\
        \hline
        OWSM-0.1B & $101$~M & $6$~layers & $6$~layers & $384$ & $6$ \\
        OWSM-1B & $1.01$~B & $18$~layers & $18$~layers & $1024$ & $16$\\
        OWLS-2B & $2.30$~B & $16$~layers & $16$~layers & $2048$ & $64$\\
        \hline\hline
    \end{tabular}
    }
\end{table}
OWSM uses an E-Branchformer~\cite{kim2023branchformer} encoder with a Transformer~\cite{vaswani2017attention} decoder, while OWLS is a standard Transformer encoder–decoder. Table \ref{tab:model_statistc} lists their full configurations. OWSM and OWLS models are trained on 180k hours of public speech data, as documented in~\cite{chen2025owls, peng2024owsm}, which do not include child speech or Flemish Dutch, although standard Dutch data are included. Therefore, domain-shifted scenarios are expected.

In our study, PEFT methods are applied to all linear layers in each model, including query, key, and value projections. For different low-rank configurations, we denote them as $LoRA_{r=rank}$, $VeRA_{r=rank}$, $DoRA_{r=rank}$, and $PiSSA_{r=rank}$. For SVFT, we follow~\cite{lingam2024svft} and use the best-performing banded matrix $\mathbf{M}$, denoted as $SVFT_{d=band~size}^B$. For SSVD, with different choices of $k$ in~(\ref{eq:ssvd-k}), rotation adaptations are applied to $p\%$ of the right singular vectors and values, denoted as $SSVD_{p=portion}$. All experiments are conducted on a single NVIDIA H100 80GB GPU.

\begin{table}[!htbp]
\centering
\small
\caption{PEFT methods across OWSM models on MyST.}
\label{tab:peft_myst}
\resizebox{0.41\textwidth}{!}{
\begin{tabular}{llrc}
\toprule
\textbf{Model} & \textbf{PEFT Method} & \textbf{\# Params} & \textbf{WER (\%) $\downarrow$}\\
\midrule
\multirow{24}{*}{\shortstack[l]{OWSM-0.1B}}
& Zero-shot & $-$ & $25.0$ \\
\cmidrule(lr){2-4}
& Full fine-tuning & $101$~M & $14.9$ \\
\cmidrule(lr){2-4}
& $LoRA_{r=8}$ & $2.07$~M & $19.2$ \\
& $LoRA_{r=16}$ & $4.13$~M & $17.3$ \\
& $LoRA_{r=32}$ & $8.26$~M & $16.4$ \\
\cmidrule(lr){2-4}
& $VeRA_{r=32}$ & $1.74$~M & $22.4 $ \\
& $VeRA_{r=64}$ & $3.36$~M & $21.1$ \\
& $VeRA_{r=128}$ & $6.59$~M & $20.2$ \\
\cmidrule(lr){2-4}
& $DoRA_{r=8}$ & $2.19$~M & $21.2$ \\
& $DoRA_{r=16}$ & $4.26$~M & $19.1$ \\
& $DoRA_{r=32}$ & $8.39$~M & $17.8$ \\
\cmidrule(lr){2-4}
& $PiSSA_{r=8}$ & $2.07$~M & $19.1$ \\
& $PiSSA_{r=16}$ & $4.13$~M & $16.8$ \\
& $PiSSA_{r=32}$ & $\mathbf{8.26}$~M & $\mathbf{16.0}$ \\
\cmidrule(lr){2-4}
& $SVFT_{d=8}^B$ & $1.23$~M & $23.9$ \\
& $SVFT_{d=16}^B$ & $2.39$~M & $20.4$ \\
& $SVFT_{d=32}^B$ & $4.67$~M & $18.2$ \\
& $SVFT_{d=64}^B$ & $9.03$~M & $16.8$ \\
\cmidrule(lr){2-4}
& $SSVD_{p=40\%}$ & $1.51$~M & $18.4$ \\
& $SSVD_{p=60\%}$ & $3.67$~M & $17.0$ \\
& $SSVD_{p=80\%}$ & $6.57$~M\textsuperscript{\dag} & $16.2$\textsuperscript{\dag} \\
\midrule
\multirow{25}{*}{\shortstack[l]{OWSM-1B}}
& Zero-shot & $-$ & $19.3$ \\
\cmidrule(lr){2-4}
& Full fine-tuning & $1.01$~B & $12.4$ \\
\cmidrule(lr){2-4}
& $LoRA_{r=8}$ & $10.56$~M & $17.6$ \\
& $LoRA_{r=16}$ & $21.13$~M & $16.1$ \\
& $LoRA_{r=32}$ & $42.26$~M & $15.1$ \\
\cmidrule(lr){2-4}
& $VeRA_{r=256}$ & $13.82$~M & $18.7$ \\
& $VeRA_{r=384}$ & $20.40$~M & $17.7$ \\
\cmidrule(lr){2-4}
& $DoRA_{r=8}$ & $11.22$~M & $16.7$ \\
& $DoRA_{r=16}$ & $21.76$~M & $15.3$ \\
& $DoRA_{r=32}$ & $42.92$~M\textsuperscript{\dag} & $14.2$\textsuperscript{\dag} \\
\cmidrule(lr){2-4}
& $PiSSA_{r=8}$ & $10.56$~M & $16.8$ \\
& $PiSSA_{r=16}$ & $21.13$~M & $15.4$ \\
& $PiSSA_{r=32}$ & $42.26$~M & $14.3$ \\
\cmidrule(lr){2-4}
& $SVFT_{d=8}^B$ & $7.00$~M & $19.0$ \\
& $SVFT_{d=16}^B$ & $13.55$~M & $16.3$ \\
& $SVFT_{d=32}^B$ & $26.52$~M & $15.1$ \\
& $SVFT_{d=64}^B$ & $51.88$~M\textsuperscript{\dag} & $14.2$\textsuperscript{\dag} \\
\cmidrule(lr){2-4}
& $SSVD_{p=22\%}$ & $9.83$~M & $15.3$ \\
& $SSVD_{p=25\%}$ & $12.50$~M & $14.5$ \\
& $SSVD_{p=29\%}$ & $16.26$~M & $14.1$ \\
& $SSVD_{p=33\%}$ & $22.16$~M & $14.1$ \\
& $SSVD_{p=40\%}$ & \textbf{31.86}~M & \textbf{13.8} \\
\midrule
\multirow{24}{*}{\shortstack[l]{OWLS-2B}}
& Zero-shot & $-$ & $20.3$ \\
\cmidrule(lr){2-4}
& Full fine-tuning & $2.30$~B & $13.1$ \\
\cmidrule(lr){2-4}
& $LoRA_{r=8}$ & $12.69$~M & $18.3$ \\
& $LoRA_{r=16}$ & $25.39$~M & $16.9$ \\
& $LoRA_{r=32}$ & $50.78$~M & $15.6$ \\
\cmidrule(lr){2-4}
& $VeRA_{r=256}$ & $14.16$~M & $19.1$ \\
& $VeRA_{r=384}$ & $20.86$~M & $18.0$ \\
\cmidrule(lr){2-4}
& $DoRA_{r=8}$ & $13.47$~M & $17.1$ \\
& $DoRA_{r=16}$ & $26.16$~M & $16.1$ \\
& $DoRA_{r=32}$ & $51.55$~M & $15.2$ \\
\cmidrule(lr){2-4}
& $PiSSA_{r=8}$ & $12.69$~M & $17.4$ \\
& $PiSSA_{r=16}$ & $25.39$~M & $16.1$ \\
& $PiSSA_{r=32}$ & $50.78$~M\textsuperscript{\dag} & $15.0$\textsuperscript{\dag} \\
\cmidrule(lr){2-4}
& $SVFT_{d=8}^B$ & $9.38$~M & $19.9$ \\
& $SVFT_{d=16}^B$ & $18.20$~M & $17.3$ \\
& $SVFT_{d=32}^B$ & $35.74$~M & $15.3$ \\
\cmidrule(lr){2-4}
& $SSVD_{p=17\%}$ & $15.04$~M & $15.1$ \\
& $SSVD_{p=20\%}$ & $21.63$~M\textsuperscript{\dag} & $14.7$\textsuperscript{\dag} \\
& $SSVD_{p=25\%}$ & $\mathbf{33.88}$~M & $\mathbf{14.6}$ \\
\bottomrule
\end{tabular}
}
\end{table}
\section{Results}
\label{sec:results}

\subsection{ASR performance}
\label{sec:results-asr}
\begin{figure*}[!htbp]
\centerline{\includegraphics[width=0.82\linewidth]{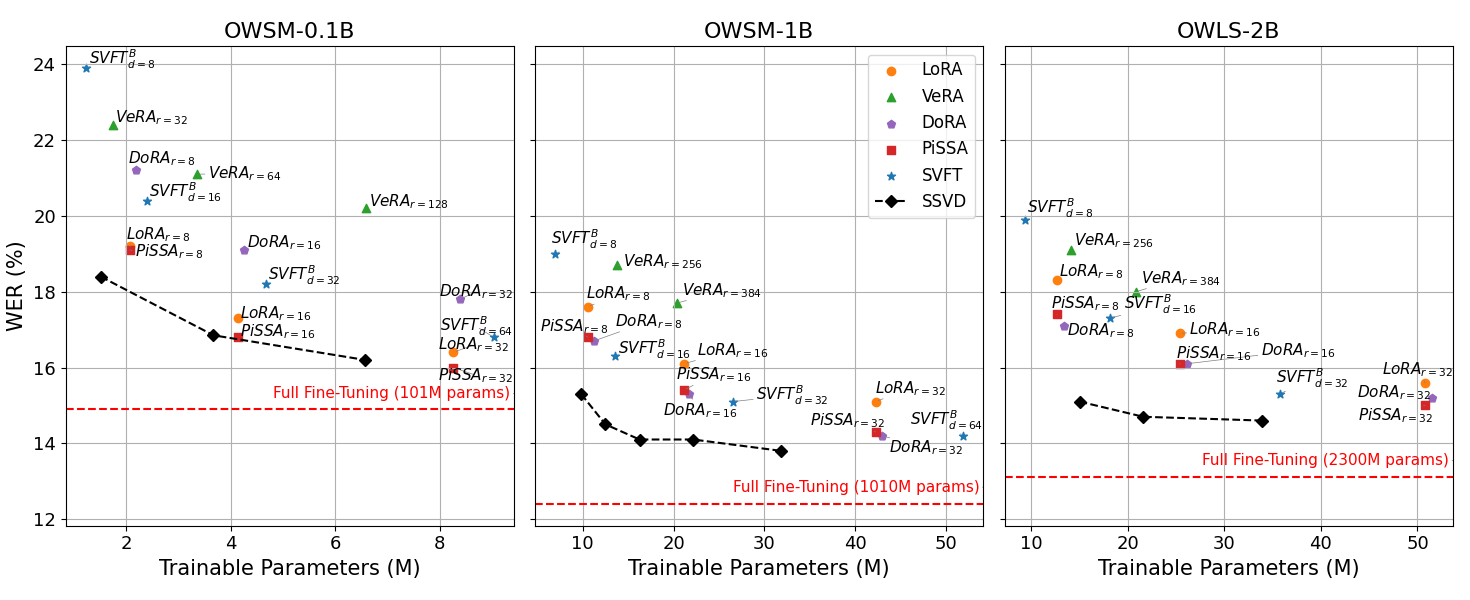}}
\caption{WER (\%) versus trainable parameters for PEFT methods fine-tuning OWSM-0.1B, OWSM-1B, OWLS-2B on MyST.}
\label{fig:trade-off}
\end{figure*}
In Table~\ref{tab:peft_myst}, we summarize the WERs on the MyST test data after fine-tuning the OWSM-0.1B, OWSM-1B and OWLS-2B models with different PEFT methods using various configurations. For each model the best-performing method is highlighted in \textbf{bold}, and the second-best is marked with a superscript \textsuperscript{\dag}.

Across all three models, the zero-shot WERs remain around 20\% or higher. After fine-tuning, for the smaller model OWSM-0.1B, all PEFT methods with similar parameter scales yield comparable performance, with $PiSSA_{r=32}$ slightly outperforming others. Notably, $SSVD_{p=80\%}$ achieves the second-best performance, with only a 0.2\% higher WER than $PiSSA_{r=32}$ while requiring nearly \textbf{2M} fewer trainable parameters (6.57M vs. 8.26M). 
\begin{figure}[!htbp]
\centerline{\includegraphics[width=0.80\linewidth]{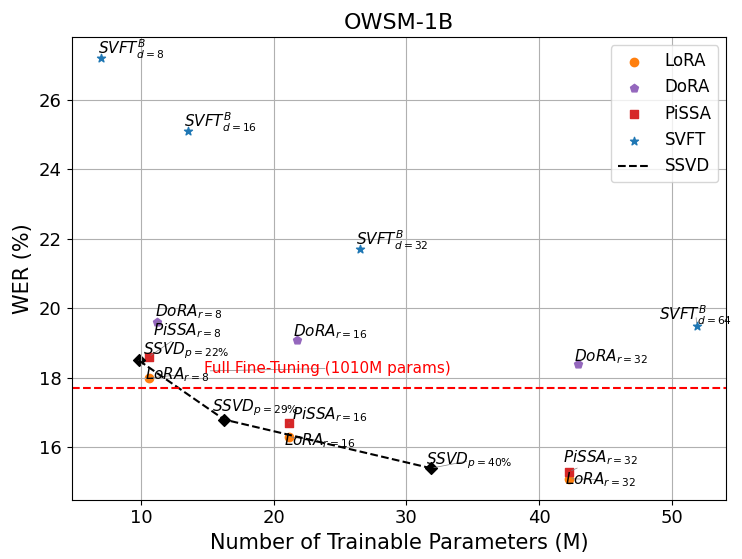}}
\caption{WER (\%) versus trainable parameters for PEFT methods fine-tuning OWSM-1B on CGN.}
\label{fig:trade-off-cgn}
\end{figure}

For the larger-scale OWSM-1B and OWLS-2B models, clearer performance trends show among PEFT methods. SSVD consistently outperforms other approaches while using significantly fewer trainable parameters. For example, when fine-tuning OWSM-1B, $SSVD_{p=40\%}$ achieves a WER of \textbf{13.8\%}, approaching the 12.4\% obtained by full fine-tuning, using only around \textbf{32M} parameters, compared to 14.2\% WER from the second best ($DoRA_{r=32}$ and $SVFT_{d=64}^B$) with approximately 52M and 43M parameters. Similarly, for the OWLS-2B model, SSVD achieves \textbf{14.7\%} WER with only \textbf{21.6M} trainable parameters, outperforming LoRA, DoRA, and PiSSA, each of which requires around 51M parameters, demonstrating SSVD's high efficiency in a domain-shift scenario. 
\begin{figure*}[!htbp]
\centerline{\includegraphics[width=0.74\linewidth]{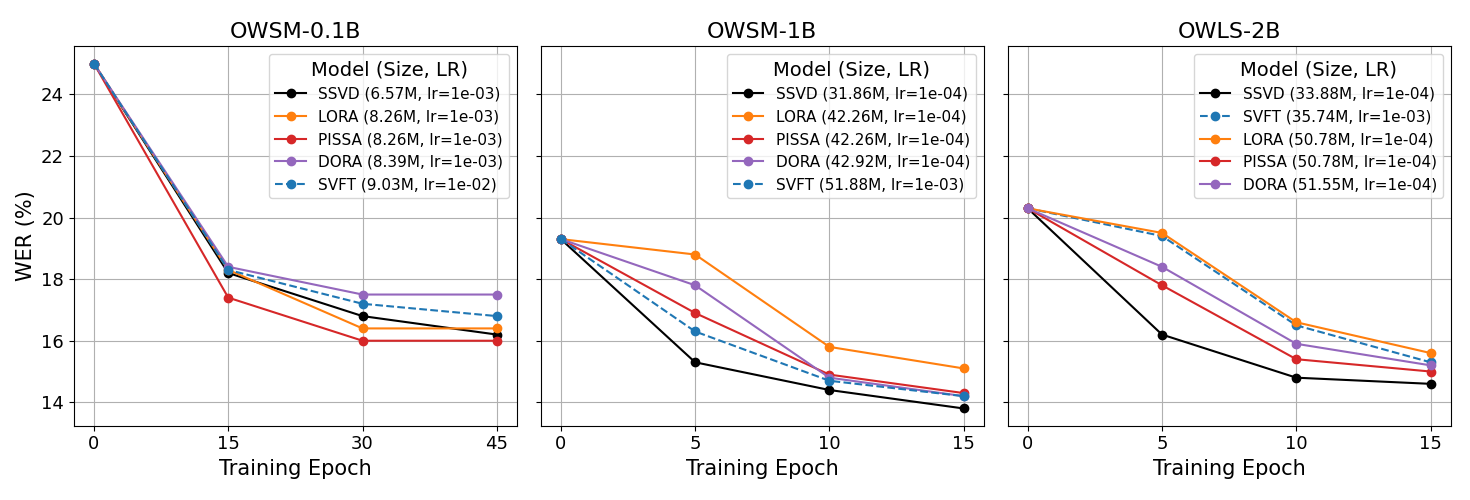}}
\caption{WER (\%) versus training epoch for PEFT methods fine-tuning OWSM-0.1B, OWSM-1B, OWLS-2B.}
\label{fig:conv_curve_myst}
\end{figure*}

To better illustrate the trade-off between ASR performance and parameter efficiency, Figure~\ref{fig:trade-off} presents WERs as a function of the number of trainable parameters for each PEFT method. Across all model scales, SSVD (depicted by black diamonds and dashed lines) consistently achieves lower WERs compared to other methods under similar trainable parameters (i.e., vertically). The performance gap becomes increasingly pronounced for larger models: in OWSM-1B and OWLS-2B, the SSVD curve lies significantly below those of other methods, indicating its strong ability to balance performance and efficiency, particularly in larger-scale ASR settings.
\begin{table}[!htbp]
\centering
\small
\caption{PEFT methods across OWSM models on CGN.}
\label{tab:peft_cgn}
\resizebox{0.392\textwidth}{!}{
\begin{tabular}{llrc}
\toprule
\textbf{Model} & \textbf{PEFT Method} & \textbf{\# Params} & \textbf{WER (\%) $\downarrow$}\\
\midrule
\multirow{22}{*}{\shortstack[l]{OWSM-0.1B}}
& Zero-shot & $-$ & $65.7$ \\
\cmidrule(lr){2-4}
& Full fine-tuning & $101$~M & $17.8$ \\
\cmidrule(lr){2-4}
& $LoRA_{r=8}$ & $2.07$~M & $25.2$ \\
& $LoRA_{r=16}$ & $4.13$~M & $21.9$ \\
& $LoRA_{r=32}$ & \textbf{8.26}~M & \textbf{19.7} \\
\cmidrule(lr){2-4}
& $DoRA_{r=8}$ & $2.19$~M & $24.2$ \\
& $DoRA_{r=16}$ & $4.26$~M & $21.9$ \\
& $DoRA_{r=32}$ & $8.39$~M & $20.7$ \\
\cmidrule(lr){2-4}
& $PiSSA_{r=8}$ & $2.07$~M & $25.5$ \\
& $PiSSA_{r=16}$ & $4.13$~M & $22.1$ \\
& $PiSSA_{r=32}$ & $8.26$~M\textsuperscript{\dag} & $20.2$\textsuperscript{\dag} \\
\cmidrule(lr){2-4}
& $SVFT_{d=8}^B$ & $1.23$~M & $36.7$ \\
& $SVFT_{d=16}^B$ & $2.39$~M & $32.6$ \\
& $SVFT_{d=32}^B$ & $4.67$~M & $28.6$ \\
& $SVFT_{d=64}^B$ & $9.03$~M & $25.0$ \\
\cmidrule(lr){2-4}
& $SSVD_{p=40\%}$ & $1.51$~M & $25.8$ \\
& $SSVD_{p=60\%}$ & $3.67$~M & $22.9$ \\
& $SSVD_{p=80\%}$ & $6.57$~M & $20.9$ \\
\midrule
\multirow{23}{*}{\shortstack[l]{OWSM-1B}}
& Zero-shot & $-$ & $46.3$ \\
\cmidrule(lr){2-4}
& Full fine-tuning & $1.01$~B & $17.7$ \\
\cmidrule(lr){2-4}
& $LoRA_{r=8}$ & $10.56$~M & $18.0$ \\
& $LoRA_{r=16}$ & $21.13$~M & $16.3$ \\
& $LoRA_{r=32}$ & $\mathbf{42.26}$~M & $\mathbf{15.1}$ \\
\cmidrule(lr){2-4}
& $DoRA_{r=8}$ & $11.22$~M & $19.6$ \\
& $DoRA_{r=16}$ & $21.76$~M & $19.1$ \\
& $DoRA_{r=32}$ & $42.92$~M & $18.4$ \\
\cmidrule(lr){2-4}
& $PiSSA_{r=8}$ & $10.56$~M & $18.6$ \\
& $PiSSA_{r=16}$ & $21.13$~M & $16.7$ \\
& $PiSSA_{r=32}$ & $42.26$~M\textsuperscript{\dag} & $15.3$\textsuperscript{\dag} \\
\cmidrule(lr){2-4}
& $SVFT_{d=8}^B$ & $7.00$~M & $27.2$ \\
& $SVFT_{d=16}^B$ & $13.55$~M & $25.1$ \\
& $SVFT_{d=32}^B$ & $26.52$~M & $21.7$ \\
& $SVFT_{d=64}^B$ & $51.88$~M & $19.5$ \\
\cmidrule(lr){2-4}
& $SSVD_{p=22\%}$ & $9.83$~M & $18.5$ \\
& $SSVD_{p=29\%}$ & $16.26$~M & $16.8$ \\
& $SSVD_{p=40\%}$ & $31.86$~M\textsuperscript{\dag} & $15.4$\textsuperscript{\dag} \\
\bottomrule
\end{tabular}
}
\end{table}

Moreover, as observed vertically in Figure~\ref{fig:trade-off}, SVD-guided methods and the directionally tuned DoRA, consistently outperform LoRA when using a similar number of trainable parameters. This trend suggests that structured initialization (as in SVD-guided methods) and directional constraints (as in DoRA) are more effective than LoRA’s random initialization with unconstrained updates, a finding that aligns with prior observations in NLP and vision tasks~\cite{liu2024dora,meng2024pissa,lingam2024svft}. 

Table~\ref{tab:peft_cgn} reports the WERs on the CGN test set after fine-tuning. Since VeRA performs poorly in this scenario, we exclude it from the table. The results for OWSM-0.1B follow a similar trend as on MyST, except SVFT, which will be discussed further in Section~\ref{sec:results-efficiency}. For the OWSM-1B model, full fine-tuning becomes less effective, and methods such as DoRA, which mimic full fine-tuning behavior, also yield suboptimal results.

SSVD achieves comparable WERs (\textbf{15.4\%} vs. 15.1\% from LoRA and 15.3\% from PiSSA) with \textbf{10M} fewer parameters (32M vs. 42M). Figure~\ref{fig:trade-off-cgn} visualizes the WER versus trainable parameter trade-off for OWSM-1B. On CGN, LoRA outperforms DoRA and SVFT at similar scales. One explanation is that CGN data require larger update deviations, and LoRA, unlike structured methods, offers more flexibility due to its unconstrained parameter updates. Nevertheless, SSVD shows a similar trend to LoRA, particularly when a larger portion of components (e.g., at 40\%) are made trainable.

\subsection{Efficiency analysis}
\label{sec:results-efficiency}
\vspace{-0.7em}
\begin{table}[!htbp]
\centering
\small
\caption{Comparing SSVD constraint methods on the MyST data.}
\label{tab:ssvd_constraint}
\resizebox{0.46\textwidth}{!}{
\begin{tabular}{ll|rcc|rc}
\toprule
\textbf{Model} & $p$ & \textbf{\# Params} & \textbf{Strict} & \textbf{Approx.} & \textbf{\# Params} & \textbf{None}\\
\midrule
\multirow{3}{*}{\shortstack[l]{OWSM-0.1B}}
& $40\%$ & $1.51$~M & $18.6$ & $18.4$ & $3.02$~M & $17.0$\\
& $60\%$ & $3.67$~M & $17.2$ & $17.0$ & $7.34$~M & $16.9$\\
& $80\%$ & $6.57$~M & $16.3$ & $16.2$ & $13.13$~M & $15.6$\\
\midrule
\multirow{5}{*}{\shortstack[l]{OWSM-1B}}
& $22\%$ & $9.83$~M & $15.1$ & $15.3$ & $19.67$~M & $14.9$\\
& $25\%$ & $12.50$~M & $14.9$ & $14.5$ & $25.00$~M & $14.6$\\
& $29\%$ & $16.26$~M & $14.5$ & $14.1$ & $32.51$~M & $13.8$\\
& $33\%$ & $22.16$~M & $14.2$ & $14.1$ & $44.32$~M & $13.7$\\
& $40\%$ & $31.86$~M & $13.8$ & $13.8$ & $63.72$~M & $13.3$\\
\midrule
\multirow{3}{*}{\shortstack[l]{OWLS-2B}}
& $17\%$ & $15.04$~M & $15.2$ & $15.1$ & $30.09$~M & $14.9$\\
& $20\%$ & $21.63$~M & $14.7$ & $14.7$ & $43.26$~M & $14.6$\\
& $25\%$ & $33.88$~M & $14.4$ & $14.6$ & $67.77$~M & $14.3$\\
\bottomrule
\end{tabular}
}
\end{table}

To further evaluate the training efficiency, Figure~\ref{fig:conv_curve_myst} presents accuracy versus epoch. Since methods with close model size have similar computing time, epoch here serves as a proxy for training time. Solid lines indicate training with a uniform learning rate, e.g., $1\mathrm{e}{-4}$, while dashed lines correspond to a tenfold higher learning rate, e.g., $1\mathrm{e}{-3}$. 
For larger-scale models, a clear convergence hierarchy shows: SSVD $>$ PiSSA $>$ DoRA $>$ LoRA. The convergence speed of SVFT, however, is highly sensitive to the chosen band size. Since SVFT updates only a fixed band of singular values, it requires a larger band size to achieve convergence comparable to PiSSA, resulting in increased trainable parameters. Additionally, SVFT benefits more from higher learning rates, as it only updates singular values while keeping singular vectors fixed, limiting the flexibility of optimization under smaller learning rates.

\subsection{Ablation study}
\label{sec:results-alba}
In Tables~\ref{tab:peft_myst} and~\ref{tab:peft_cgn}, SSVD is implemented in the ``Approx.'' variant as described in Section~\ref{sec:method-approximate}, which introduces minor orthogonality errors. As further evaluated in Table~\ref{tab:ssvd_constraint}, we compare this variant against the ``Strict'' implementation (Section~\ref{sec:method-strict}) and a ``None'' constraint baseline (Section~\ref{sec:method-none}). The results show that ``Strict'' and ``Approx.'' yield nearly identical WERs, suggesting that the orthogonality deviations in the approximate implementation are relatively minor and do not significantly affect performance in the PEFT setting. The ``None'' constraint implementation, as explained in Section~\ref{sec:method-none}, doubles the number of trainable parameters. While increasing parameters can improve performance, the ``None'' constraint performs worse than ``Strict'' and ``Approx.'' when the number of trainable parameters is close. For instance, with OWLS-2B, it achieves a WER of 14.9\% using 30.09M parameters, higher than the 14.7\% WER achieved with only 21.63M parameters.

\section{Conclusion}
In this work, we presented a comprehensive evaluation of PEFT methods for adapting large-scale speech foundation models under domain-shifted conditions, such as child speech and dialectal variation. We integrated and benchmarked SoTA PEFT techniques, including LoRA, DoRA, VeRA, PiSSA, SVFT, and our proposed SSVD, within ESPnet across model sizes from 0.1B to 2B. Our results highlight that while several PEFT methods perform comparably on small models, performance gaps widen at larger scales, especially under constrained compute budgets. Among all methods, SSVD consistently achieves the best trade-off between performance and parameter efficiency, closely approaching full fine-tuning performance with significantly fewer trainable parameters. 



\section*{Acknowledgment}
Experiments of this work used the Bridges2 system at PSC and Delta system at NCSA through allocations CIS210014 and IRI120008P from the Advanced Cyberinfrastructure Coordination Ecosystem: Services \& Support (ACCESS) program, supported by National Science Foundation grants \#2138259,\#:2138286, \#:2138307, \#:2137603, and \#:2138296. This research was supported by the Flemish Government under ``Onderzoeksprogramma AI Vlaanderen'', the FWO-SBO grant S004923N: NELF, and the FWO grant V401325N.
\printbibliography

\end{document}